\documentclass[10pt,twocolumn,letterpaper]{article}
\usepackage[accsupp]{axessibility} 
\usepackage{wacv}
\usepackage{times}
\usepackage{graphicx}
\usepackage{amsmath}
\usepackage{amssymb}
\usepackage[utf8]{inputenc} 
\usepackage[T1]{fontenc}    
\usepackage{url}            
\usepackage{booktabs}       
\usepackage{amsfonts}       
\usepackage{nicefrac}       
\usepackage{microtype}      
\usepackage{xcolor}         
\usepackage{mathtools}
\usepackage{subcaption}
\usepackage{graphics} 
\usepackage{mathptmx} 
\usepackage{multirow}
\usepackage{tabularx,wrapfig,lipsum}

\def\wacvPaperID{527} 

\wacvfinalcopy 

\ifwacvfinal
\fi

\usepackage[breaklinks=true,bookmarks=false]{hyperref}

\ifwacvfinal
\fi

\begin{document}

\title{LwPosr: Lightweight Efficient Fine Grained Head Pose Estimation}

\author{Naina Dhingra\\
ETH Zurich\\
{\tt\small ndhingra@ethz.ch}}


\maketitle

  \begin{abstract}
This paper presents a lightweight network for head pose estimation (HPE) task. While previous approaches rely on convolutional neural networks, the proposed network \textit{LwPosr} uses mixture of depthwise separable convolutional (DSC) and transformer encoder layers which are structured in two streams and three stages to provide fine-grained regression for predicting head poses. The quantitative and qualitative demonstration is provided to show that the proposed network is able to learn head poses efficiently while using less parameter space. Extensive ablations are conducted using three open-source datasets namely 300W-LP, AFLW2000, and BIWI datasets. To our knowledge, (1) \textit{LwPosr} is the lightest network proposed for estimating head poses compared to both keypoints-based and keypoints-free approaches; (2) it sets a benchmark for both overperforming the previous lightweight network on mean absolute error and on reducing number of parameters; (3) it is first of its kind to use mixture of DSCs and transformer encoders for HPE. This approach is suitable for mobile devices which require lightweight networks. 
\end{abstract}

\section{Introduction}
\vspace{-2mm}
Analysis of facial gestures and landmarks have been actively researched since past few decades \cite{bulat2017far,kumar2017kepler}. Some of the sub-topics involving facial analysis are face recognition, face detection and alignment, head pose estimation (HPE), facial emotion recognition, etc. HPE is crucial for non-verbal communication (NVC) in helping blind people, advanced driver assistance systems, behaviour analysis, in-person attention modeling, human-computer interaction, virtual/augmented reality, etc.

Depth images are used in \cite{martin2014real,mukherjee2015deep} to predict head poses since estimating yaw, pitch, and roll angles are a 3D vision problem. Using depth images, it is possible to achieve high prediction performance but they require RGB-D cameras. Past works on HPE have also used facial landmark for predicting 3D vectors \cite{kazemi2014one,bulat2017far}. But it leads to expensive computation costs and larger models. Therefore, these techniques are not suitable to be utilized on platforms which require low computation power and have limited memory. This calls for designing lightweight networks as they are serviceable for employing them on mobile devices such as drones, movable robots, mobile phones, advanced driver assistance systems. 

Using holistic approach of predicting 3D head pose from single image has shown to perform better than the keypoints-based methods \cite{ruiz2018fine}. Keypoints-based techniques are required to have good detection accuracy of keypoints. Furthermore, it makes the network unnecessarily complex. Also, its performance depends on the standard of the 3D head model which is required for establishing correspondences with keypoints. It is common to jointly predict facial landmarks and head poses but the primary goal in these works is generally to predict the landmarks \cite{kumar2017kepler,ranjan2017hyperface,ranjan2017all}. So, being the secondary task, predicted head poses are not very accurate. In this work, a complete end-to-end learning based network is designed for HPE keeping in mind to use less number of parameters and at the same time to have a good performance.

Transformers \cite{vaswani2017attention} have recently become famous in computer vision problems considering their good performance for natural language processing applications. They have been used in image classification \cite{dosovitskiy2020image}, image enhancement \cite{yang2020learning}, image segmentation \cite{wang2020end}, and object detection \cite{carion2020end}. They are being preferred for various different modalities. The positive point about their use is that they help in reducing the inductive biases that come into designing the networks. They are also shown to get an enormous parameter number reduction while maintaining the estimation performance \cite{dosovitskiy2020image}. Attention in the form of self-attention mechanism is the main building basis of transformers. It also makes it possible to learn the alignments and order which exist inside the data  \cite{vaswani2017attention,bahdanau2014neural}. Hence, it helps to remove the requirement for watchfully designing the inductive biases to a large extent. To our best knowledge, this is the first work using transformer encoders in a network with depthwise separable convolutions (DSCs) for designing a lightweight network for HPE.

The main contributions of this paper are given as:
\begin{itemize}
    \item A novel architecture is proposed based on fine-grained regression using DSCs and transformer encoders. This network is able to predict head pose given by roll, yaw, and pitch angles. It consists of two streams having three stages each. Each stage consists of DSCs and/or transformer encoder and each stage has its own prediction of head poses. The weighted mean of the three head pose estimations is the final head pose prediction. \vspace{-1mm}
    \item The proposed \textit{LwPosr} is compared with the state-of-the-art methods based on its performance and computational     parameters.\vspace{-1mm}
    \item  Qualitative results are illustrated to compare them with the ground truth and with FSA-Net \cite{yang2019fsa}. Quantitative results are also visualized on a log-scale plot.\vspace{-1mm}
    \item  Extensive ablation study is presented to understand the nuances of varying several parameters in the network.\vspace{-1mm}
    \item  It is the lightest network proposed for HPE to our best knowledge. The proposed network outperforms the state-of-the-art methods which have comparable number of parameters. This work will promote the use of transformer encoders for future research in designing the algorithms for HPE with less number of parameters. This work will also have an impact on the research in similar domains to employ \textit{LwPosr} variants which can have good estimations using less number of parameters.\vspace{-1mm}
\end{itemize}

\section{Related Work}
\subsection{Head Pose Estimation}
Head pose estimation can be solved in two ways. One way is to use the facial landmarks (or keypoints) along with a reference 3D head model to predict the head pose from these landmarks. The other way is to utilize the complete face appearance to predict the pose using either direct relation from image to pose or using a appearance face model. 

\textbf{Keypoints-Based Approaches:} 
In these approaches, keypoints are estimated initially and then utilized for predicting the head pose. 3D vision techniques are used as in \cite{dementhon1995model} where first 2D face keypoints are detected. Some techniques \cite{cao2014face,xiong2013supervised,xiong2015global,lee2015face,dollar2010cascaded} sketch rough faces and then modify these faces incrementally to align them with the real faces using regression. Model-based techniques generate human face models using landmarks and generate correspondences on real faces using learned appearance models. In \cite{guo2020towards}, an optimization technique to regress 3DMM parameters is utilized and prediction of rotation matrix is performed using a network. In \cite{kumar2017kepler}, a heatmap-CNN is defined which refines the facial landmarks iteratively and HPE is a side product of the total task. Techniques employing deep learning to predict 3D face models using trained network perform better than the traditional approaches. These keypoint based techniques under-perform as compared to non-keypoints based approaches.

\textbf{Non-keypoints-Based Approaches:} 
Non-keypoints based methods have the benefit of not requiring the acquisition of the landmarks which makes the task less intensive. Different approaches have used different kind of modalities to solve HPE. Many recent approaches have employed a single RGB image for estimating the head poses \cite{chang2017faceposenet,ranjan2017all,yang2019fsa,ruiz2018fine,ranjan2017hyperface,kumar2017kepler,zhou2020whenet,zhang2020fdn}. In \cite{ruiz2018fine}, a ResNet50 is used along with multi-loss technique where loss consists of binned pose classification and regression for all three angles separately. In \cite{yang2019fsa}, a stage-wise regression is performed using a CNN based network. A new feature aggregation approach with attention mechanism is used to combine features from each stage. In \cite{zhou2020whenet}, multi-loss approach is used to work on full range of yaws. In \cite{zhang2020fdn}, a feature decoupling model is used to learn discriminative feature for individual pose angle using a cross-category loss. In \cite{cao2021vector}, rotation matrix is used to represent three vectors for depicting the head pose. They used three sub-networks which are very similar to that used in \cite{yang2019fsa} to estimate the three vectors, i.e., total nine values instead of three angles. In this paper, non-keypoints-based technique is used as it is shown to have several advantages over the keypoints-based techniques.

\subsection{Transformers}
Transformers have recently dominated the natural language modeling using their self-attention mechanism which has shown to have the ability to capture the global information \cite{vaswani2017attention,devlin2018bert,radford2018improving}. They have been recently used for computer vision tasks such as image classification \cite{dosovitskiy2020image,touvron2020training}, image generation \cite{parmar2018image}, image enhancement \cite{yang2020learning,chen2020pre}, image segmentation \cite{wang2020max,wang2020end}, etc. Various different versions of vision transformers have been proposed and shown to perform well on large data (such as on ImageNet-22k, JFT-300M) \cite{wang2021pyramid,wu2021centroid,yuan2021tokens,touvron2020training}. In \cite{graham2021levit}, a hybrid neural network is developed for fast image classification where a mixture of convolution and transformer is used with new attention bias. In \cite{wu2021cvt}, a new network based on a mixture of convolutions and vision transformer is designed to have the positives of both convolutions and transformer. In \cite{stoffl2021end} also, a mixture of CNN with a transformer is utilized for multi-instance pose estimation.  Inspired by the these works, this work employs mixture of convolutions and transformer to get best characteristics of both for HPE.

\section{Method}
\label{sec:method}

\begin{figure*}[htbp!]
\setlength\abovecaptionskip{-0\baselineskip}
\setlength\belowcaptionskip{2pt}
\begin{center}
\includegraphics[height=6cm, width=13.5cm]{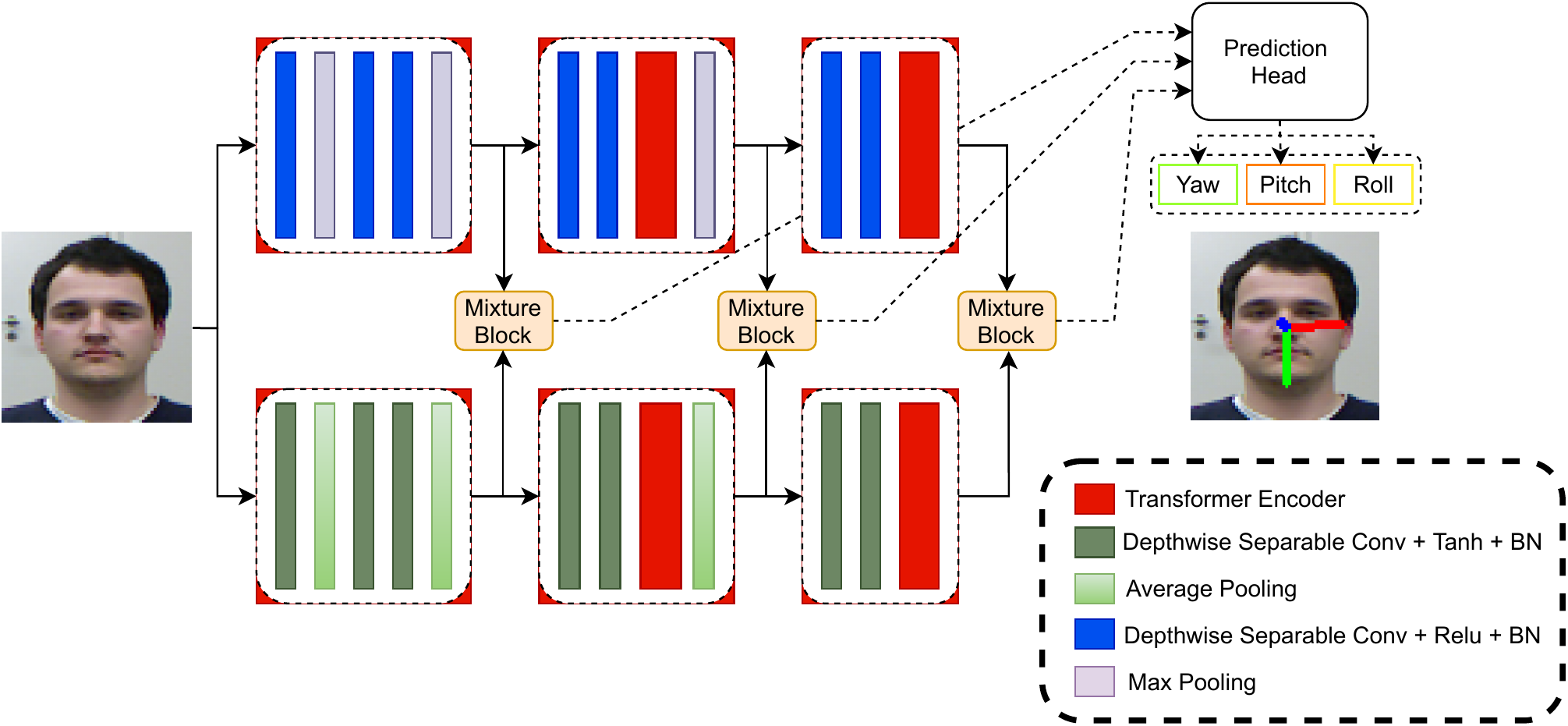}
\caption{Architecture of \textit{LwPosr}. It consists of two streams, each stream has three stages.  Each of the three stage contributes to the estimation of pose directly in a fine-grained fashion. Mixture Block consists of element-wise multiplication of the output from two streams. Prediction head consists of $1 \times 1$ convolutional layer and a fully connected layer which regresses three head pose predictions from three stages. The weighted mean of all the three predictions is calculated to get final head pose estimation, i.e., yaw, pitch, and roll. The network is trained end-to-end.}
  \label{LwPosr}
\end{center}
\end{figure*}

LwPosr consists of two streams and $Q$ stages, where $Q = 3$ as shown in Figure \ref{LwPosr}. Each of the two streams extract a feature map at $Q^{th}$ stage. The outputs from both the streams are fused together by element-wise multiplication followed by $1\times1$ convolutional layer (conv) to transform the features to c channels. The output of all the stages have same dimensions. The individual stage output is fed into a linear layer to predict the pose. The predicted head poses $p_q$'s are combined using Eq. \ref{eq:combine}, in which $P$ is the final predicted head pose and $p_q$ is the $q^{th}$ predicted pose. The weighted term ${\alpha}_q$ is found empirically and is shown in section \ref{sec:exp} ablation study. The problem of HPE has a regression formulation. So, mean absolute error (MAE) is used as a loss function and evaluation metric similar to used in \cite{yang2019fsa}.

\begin{equation} \label{eq:combine}
    P= \Sigma_{q=1}^{Q} {\alpha}_q p_q 
\end{equation}

\subsection{Single Stage Architecture:}
Each of the stage consists of stacked depthwise separable filters (DSF). The technique of replacing standard convolutional filter with two consecutive layers, i.e., with depthwise and pointwise convolution filters for mobile networks is adapted from \cite{howard2017mobilenets,yang2019fsa}. DSC is a structured convolutional layer which factorizes the conventional convolutional layer into a depthwise and a pointwise (also known as $1 \times 1$ conv) convolutional layer. In a conventional convolution, filtering and combining of input to output features takes place in a single step. DSC breaks this single step into two individual layers, i.e., a layer for filtrating and a second layer for combining. This splitting step helps in considerably reducing the computation power required and further aids in reducing the model size. 
The DSC is followed by a transformer encoder. The convolution layer helps in capturing the local and global dependencies, and transformer encoder aids in understanding the global information. In \cite{stoffl2021end} and \cite{yang2020transpose}, transformer encoder is used with convolutional based networks to understand the image-based features. Motivated by these works, in this paper, the output from convolutional layer is converted in the form of sequential features to use as an input to transformer encoder. The description of the functioning and decreased dimension of the transformer encoder is presented in section~\ref{sec:transformer_encoder}.

\subsection{Reducing Number of Parameters using Depthwise Separable Filters:}
In a standard convolutional layer, $F_m \times F_n \times C_{in}$ are given as input I to the layer, where $F_m \times F_n$ is the spatial size of the input feature map, $C_{in}$ is the input channel size. Output of this convolution layer has $F'_m \times F'_n \times C_o$, where $F'_m \times F'_n$ is spatial feature size and $C_o$ is the output channel size. The convolutional kernel $\hat{\mathbf{K}}$ in standard convolutional layer is parameterized by $F_k \times F_k \times C_{in} \times C_o$ where $F_k \times F_k$ is the spatial size of the kernel, and $C_i$ is input channel size and $C_o$ is output channel size. The output of a convolutional layer with stride one and padding is given by Eq. \ref{eq:conv}.

\begin{equation} \label{eq:conv}
\hat{\mathbf{\mathcal{F}} }_{x,y,n} = \sum_{i,j,m} \hat{\mathbf{K}}_{i,j,m,n} \cdot \mathbf{I}_{x+i-1,y+j-1,m}
\end{equation}

The computational cost of this layer as given in Eq. \ref{eq:computation_conv} depends on the kernel size $F_k \times F_k$, dimension of input and output channels, i.e., $C_{in}$ and $C_o$, and $F_m \times F_n$.
\begin{equation} \label{eq:computation_conv}
{Comp Cost}_{\text{conv}} = F_k \cdot F_k  \cdot C_{in} \cdot C_o \cdot F_m \cdot F_n
\end{equation}

DSC consists of two separate layers: (1) depthwise convolution to apply a single filter for each input depth or channel; (2) pointwise convolution to create a linear combination of the features generated by the depthwise convolutional layer. Unlike \cite{howard2017mobilenets}, in this work, batchnorm and relu are used only after the pointwise convolution layer.

The output of a depthwise convolution with one filter for each input channel is given by Eq. \ref{eq:depth_sep_conv}. The ${\mathbf{K}}$ kernel for depthwise convolutional layer has size $F_k \times F_k \times C_{in}$. The $n^{th}$ filter of the kernel $\hat{\mathbf{K}}$ is used on the $n^{th}$ channel to give $n^{th}$ output feature map.
\begin{equation} \label{eq:depth_sep_conv}
{\mathbf{\mathcal{F}}}_{x,y,n} = \sum_{i,j} \mathbf{K}_{i,j,n} \cdot \mathbf{I}_{x+i-1,y+j-1,n}
\end{equation}

The corresponding computational cost for the depthwise convolutional (DC) layer is .

\begin{equation} \label{eq:cost_depthwise}
{Comp Cost}_{\text{DC}} = F_k \cdot F_k \cdot C_{in} \cdot F_m \cdot F_n
\end{equation}

This depthwise convolution provides with the filtration of the input channels. But the combination of these filtered output does not take place by the  depthwise convolutional layer. Therefore, a pointwise convolutional layer is used to combine them. It linearly combines the output features from the depthwise convolutional layer.
Both the layers, i.e., depthwise and pointwise convolution form a DSC. The computation cost of the DSC is given by Eq. \ref{eq:computation_depthwisesep}
\begin{equation}\label{eq:computation_depthwisesep}
{Comp Cost}_{\text{DSC}}=F_k \cdot F_k \cdot C_in \cdot F_m \cdot F_n +  C_{in} \cdot C_o \cdot F_m \cdot F_n\end{equation}

By comparing both the computational costs given in Eq.  \ref{eq:computation_conv} and \ref{eq:computation_depthwisesep}, it results into Eq.~\ref{eq:comparison_costs} which shows that DSC have lower number of parameters than conventional convolutional layer.
\begin{equation}\label{eq:comparison_costs}
\frac{F_k \cdot F_k \cdot C_{in} \cdot F_m \cdot F_n +  C_{in} \cdot C_{o} \cdot F_m \cdot F_n}{F_k \cdot F_k \cdot C_{in} \cdot C_{o} \cdot F_m \cdot F_n} =\frac{1}{C_{o}} + \frac{1}{F_k^2}
\end{equation}

This work also uses $3 \times 3$ DSCs which use approximately 8 to 9 times lower computational power than conventional convolutional layer as shown in \cite{howard2017mobilenets}.

\subsection{Transformer Encoder:}\label{sec:transformer_encoder}
Each stage structure except the first one has a transformer encoder, i.e., second and third stage in both the streams. Transformer encoder follows a very similar structure as proposed in \cite{vaswani2017attention}. To reform the output from the DSC, a reshaper is used. It converts the output features into a sequential form which can be used as an input to transformer encoder. The reshaper performs this operation by multiplying the height and width of the features and then performs permutation on them. Consider the output features from DSC layers $\mathbf{\mathcal{F}} \in \mathbb{R}^{B \times C \times H \times W}$, where $B$ is the batch-size, $C$ is the number of channels, $H$ stands for the height, and $W$ for the width. Reshaper transforms it to $\mathbf{\mathcal{F}} \in \mathbb{R}^{B \times A \times C}$, where $A = H \times W$. This output $\mathbf{\mathcal{F}}$ is in sequential form and is apt to be fed into a transformer encoder.

Transformer encoder uses multi-head self attention mechanism. It gets a sequential input $\mathbf{\mathcal{F}}\in \mathbb{R}^{B \times A \times C}$. The inputs are used to calculate values $V  \in \mathbb{R}^{A \times C}$, keys $K  \in \mathbb{R}^{A \times C}$, and queries $Q  \in \mathbb{R}^{A \times C}$ as shown in Eq. \ref{eq:keys}.

\begin{equation}\label{eq:keys}
    Q_{h} = \mathbf{\mathcal{F}} * W^{Q}_{h} \text{ and }  K_{h} = \mathbf{\mathcal{F}} * W^{K}_{h} \text{ and } V_{h} = \mathbf{\mathcal{F}} * W^{V}_{h}
\end{equation}

Here, $h$ represents used attention head, and $W^{Q}_{h} \in \mathbb{R}^{C \times C_{k}}$, $W^{K}_{h} \in \mathbb{R}^{C \times C_{k}}$, and $W^{V}_{h} \in \mathbb{R}^{C \times C_{v}}$ represent parameter matrices. In this work, position embeddings are used in transformer encoder. Without position embeddings, transformer encoder would be permutation-invariant, hence would not consider the spatial formation of the features. The output $Z_{i}$ of the scaled dot-product attention head is given by Eq \ref{eq:attention}
\begin{equation} \label{eq:attention}
    Z_{h} = softmax\left(\frac{Q_{h} * K_{h}^{T}}{\sqrt{C_{k}}}\right)
\end{equation}
The output  $Z_{h}$ from attention heads are concatenated to form \begin{equation}
    Z = concatenate(Z_{1}, Z_{2}, ... , Z_{n_{heads}}).
\end{equation}

The Add \& Norm layer is used for adding the residual connection of the input with the output of previous multi-head attention and for normalization by calculating the mean and variance of the cumulative inputs using the technique given in \cite{ba2016layer}. It also alleviates the problem of `covariate shift'. This is followed by a feed-forward layer and activation function (see Section \ref{subsec:ablation} for results on two different activation functions). It is followed by another Add \& Norm layer. For detailed working of transformer, refer to \cite{vaswani2017attention}.

\subsection{Architecture Details:}  There are two streams in the network as used in \cite{yang2017deepcd,yang2018ssr,yang2019fsa}. Similar to \cite{yang2019fsa}, two basic DSC blocks are used:
${DSC}_r(c) \equiv [DSC(3 \times 3, c)-BN-ReLU] $ and ${DSC}_t (c) \equiv [DSC(3 \times 3, c)-BN-Tanh] $, where c is a parameter and BN stands for batch-normalization. $TransEnc$ denotes transformer encoder with 8 heads, 3 encoder layers, 32 number of expected features in the encoder (denoted as k), 64 feedforward dimension. The first stream has $[{DSC}_r(16)-AvgPool(2 \times 2)-{DSC}_r(32)-
{DSC}_r(32)-AvgPool(2 \times 2)] - [{DSC}_r(32)-{DSC}_r(32)-TransEnc-AvgPool(2 \times 2)]
- [{DSC}_r(32)-{DSC}_r(32)-TransEnc]$. A stage is formed by each pair of square brackets. The second stream has same structure as the first stream except that it uses  ${DSC}_t$ instead of  ${DSC}_r$ and $MaxPool$ instead of $AvgPool$.
Mixture block consists of element-wise multiplication of the features from the two streams. Prediction head consists of $1 \times 1$ convolutional layer which lowers the number of channels from 32 to 16 (first stage has AvgPool after $1 \times 1$ conv). The flattened array is then fed into linear layer having three output units, one each for yaw, pitch, and roll. The final head pose is calculated using the three predicted head poses (one from each stage) by taking their weighted mean as given in Eq. \ref{eq:combine}.

\vspace{-2mm}
\section{Experiments} \label{sec:exp}
\vspace{-2mm}
This section describes the used datasets, implementation details, evaluation protocols, quantitative and qualitative results, comparison with the state-of-the-art, and extensive ablation study.

\subsection{Datasets} \label{sec:dataset}
Three open-source popular datasets are adopted in this work, i.e., the 300W-LP~\cite{zhu2016face}, AFLW2000~\cite{zhu2015high}, and BIWI~\cite{fanelli2013random} datasets. 

\textbf{300W-LP Dataset:}
It is acquired from the 300W dataset \cite{sagonas2013300} which is an amalgamation of several sub-datasets. Some of the examples for sub-datasets are: AFW \cite{zhu2012face}, LFPW \cite{belhumeur2013localizing}, HELEN \cite{zhou2013extensive}, IBUG \cite{sagonas2013300}, etc. Face profiling is used with 3D meshing to expand 300W dataset to produce 61,225 image samples for large poses \cite{zhu2016face}. Flipping is also used to further expand the dataset images to 122,450 synthesized image samples. Therefore, the resulting synthesized dataset by \cite{zhu2016face} has its name as the 300W across large poses (300W-LP). 

\textbf{AFLW2000 Dataset:}
It has first 2,000 images present in AFLW \cite{koestinger2011annotated} dataset. It includes ground-truth 3D faces along with 68 landmarks for each face. These 3D faces have a large variations in their illuminations, expressions, and poses.

\textbf{BIWI Dataset:}
It has total of 15,678 frames generated out of 24 videos having 20 different subjects. These videos are taken in the controlled indoor environment. This dataset does not inculcate human heads bounding boxes. So, in this work, the human faces are detected and cropped to get the bounding box of the face using MTCNN \cite{zhang2016joint} similar to \cite{cao2021vector,yang2019fsa}.

\subsection{Experiment Protocols} \label{sec:protocols}
\textbf{Protocol} \textit{P1}: Similar settings are followed as in \cite{yang2019fsa,doosti2020hope,cao2021vector} to have fair comparison of the results. Training is performed on synthetic 300W-LP dataset. Testing is performed on real-world datasets: (1) AFLW2000, (2) BIWI datasets. Similar to \cite{yang2019fsa,doosti2020hope,cao2021vector}, this protocol does not utilize tracking and uses rotation angles between [-99\textdegree, 99\textdegree] with MTCNN \cite{zhang2016joint} face detection algorithm. 

\textbf{Protocol} \textit{P2}: It uses BIWI dataset which is divided into 70\% training and 30\% testing dataset, same as performed by \cite{cao2021vector,yang2019fsa}. MTCNN is used to detect the faces with empirical tracking method so that all the faces get detected successfully. This protocol has been used with several different modalities in the literature. In this work, single RGB image is used.

\subsection{Implementation Details:}
 Pytorch library is used for the implementation of the LwPosr. The implementation parameters are kept same as described by \cite{yang2019fsa,cao2021vector} so that the comparison of results is fair. During training, random cropping and scaling is used by a factor of 0.8-1.2. The training is carried on for 90 epochs using an Adam optimizer and with initial learning rate of 0.001. This learning rate is reduced after 30 steps by a factor of 0.1. Batch-size used is 16 images for protocol $P_1$ and 8 images for protocol $P_2$. The experimental implementation was carried on an GeForce RTX 2080 Ti GPU (training time is ~8 hours).

\subsection{Quantitative Results} \label{sec:results}

\begin{table*}[htbp!]
\small
\centering
\caption{Results for Protocol $P_1$ trained on 300W-LP dataset, tested on BIWI \cite{fanelli2013random} (left) and AFLW2000 \cite{zhu2016face} (right) dataset}
\label{Tab:result_300wlp}
\begin{tabular}{l|l|llll|llll}
\hline
        & & \multicolumn{4}{c}{BIWI} & \multicolumn{4}{|c}{AFLW2000}\\
\hline
Method   & Param $10^6$ & Yaw  & Pitch & Roll & MAE  & Yaw   & Pitch  & Roll & MAE  \\
\hline
3DDFA~\cite{zhu2016face}  &- & 36.20 & 12.30 & 8.78  & 19.10 & 5.40  & 8.53  & 8.25  & 7.39  \\
KEPLER~\cite{kumar2017kepler} &- & 8.80  & 17.3 & 16.2 & 13.9 & -& - & -&-\\
Dlib (68 points)~\cite{kazemi2014one} & 6-24 & 16.8 & 13.8 & 6.19  & 12.2 & 23.1 & 13.6 & 10.5 & 15.8\\
FAN (12 points)~\cite{bulat2017far}  &  $\sim$36.6  & 6.36  & 12.3 & 8.71 & 9.12    & 8.53  & 7.48  & 7.63  & 7.88  \\
Hopenet(a = 1)~\cite{ruiz2018fine}  & 23.9 & 4.81  & 6.61  & 3.27  & 4.90  & 6.92  & 6.64  & 5.67  & 6.41  \\
Hopenet(a = 2)~\cite{ruiz2018fine} & 23.9 & 5.12  & 6.98  & 3.39  & 5.18  & 6.47  & 6.56  & 5.44  & 6.16  \\
Shao ~\cite{shao2019improving} &24.6 & 4.59 & 7.25 & 6.15  & 6.00  & 5.07  & 6.37 & 4.99 & 5.48\\
SSR-Net-MD ~\cite{yang2018ssr}   &0.2 & 4.49  & 6.31  & 3.61  & 4.65  & 5.14  & 7.09  & 5.89  & 6.01  \\
FSA-Caps-Fusion~\cite{yang2019fsa}& 1.2 & 4.27  & 4.96  & 2.76  & 4.00  & 4.50  & 6.08  & 4.64  & 5.07  \\
TriNet~\cite{cao2021vector} &  $\sim$26 &  4.11 &  4.76 & 3.05  & 3.97 & 4.04  & 5.78 & 4.20 & 4.67 \\
WHENet \cite{zhou2020whenet} & 4.4 & 3.99  & 4.39  & 3.06  & 3.81 & 5.11  & 6.24  & 4.92  & 5.42 \\
EVA-GCN \cite{xin2021eva} & $\sim$3.3 & 4.46 & 5.34 & 4.11 & 4.64 &  4.01 & 4.78 & 2.98 & 3.92 \\
\hline

\textit{LwPosr} & \textbf{0.15} & 4.11 & 4.87 & 3.19 & 4.05 & 4.80 & 6.38 &  4.88 & 5.35 \\
\textit{LwPosr} $\alpha$ & \textbf{0.15} & 4.41 & 5.11 & 3.24 & 4.25 & 4.44 & 6.06 &  4.35 & 4.95 \\

\end{tabular}

\label{tab:narrow}
\end{table*}

\textit{LwPosr} is compared to existing methods (with both lightweight and heavier techniques) for HPE on open-source benchmark datasets. Table \ref{Tab:result_300wlp} and \ref{Tab:results_biwi} illustrate the comparison of the number of parameters and MAE for the compared frameworks. They are as follows. (1) 3DDFA~\cite{zhu2016face} uses a cascaded CNN network which tends to fit 3D model to a single image. It is described to perform well on occlusions. (2) KEPLER~\cite{kumar2017kepler} uses Heatmap-CNN to understand the local and global features to detect keypoints for face alignment and HPE is a by-product task. (3) Dlib~\cite{kazemi2014one} is a standard well used library for faces for applications of HPE, face detection, keypoints prediction. (4) FAN~\cite{bulat2017far} utilizes landmark localization network combined with residual block forming a multi-stage network with multi-scale features. (5) Hopenet~\cite{ruiz2018fine} is ResNet based fine-grained structure using classification and regression loss. (6) Shao ~\cite{shao2019improving} works on the adjustment of margins of detected bounding box for faces using a CNN architecture. (7) SSR-Net-MD~\cite{yang2018ssr} uses soft stage-wise regression. (8) FSA-Caps-Fusion~\cite{yang2019fsa} learns feature aggregation for fine-grained regression. (9) TriNet~\cite{cao2021vector} uses ResNet along with feaure agregagtion as used in ~\cite{yang2019fsa} for three different streams. It is a memory intensive network. (10) WHENet \cite{zhou2020whenet} is also one of the low memory networks (having $4.4 \times 10^6$ number of parameters) which uses efficientNet (lightweight backbone) and uses multi loss (classification and regression loss). (11) VGG16~\cite{gu2017dynamic} and VGG16+RNN~\cite{gu2017dynamic} are highly memory expensive models with CNN and CNN with RNN respectively, using Bayesian filters analysis. (12) DeepHeadPose~\cite{mukherjee2015deep} uses depth images (low resolution) and uses regression and classification for each angle prediction. (13) Martin~\cite{martin2014real} too uses depth images and performs registration to a 3D head model.

\textbf{Protocol $\boldsymbol{P_1}$:} To our best knowledge, \textit{LwPosr} is the lowest memory requiring network with just $0.159 \times 10^6$ parameters. TriNet has lowest MAE on AFLW2000 dataset with a difference of 0.68 on MAE but it has $\sim163.5$ times more number of parameters. WHENet has least MAE on BIWI dataset with a difference of 0.24 but it requires $27.7$ times more number of parameters than \textit{LwPosr}. In the literature, SSR-Net-MD has the least number of parameters in the network. \textit{LwPosr} has outperformed it on MAE on both the datasets by 0.6 on BIWI and 0.66 on AFLW2000 dataset simultaneously reducing the number of parameters. Hence, \textit{LwPosr} has significantly improved the benchmark for networks with least number of parameters for HPE to our best knowledge and as is also evident in Table \ref{Tab:result_300wlp}. \textit{LwPosr} $\alpha$ has performed better than \textit{LwPosr} on AFLW2000 dataset. The only difference is that \textit{LwPosr} $\alpha$ uses Gelu activation function in transformer encoders instead of Relu.

\textbf{Protocol $\boldsymbol{P_2}$:} It is evident from Table \ref{Tab:results_biwi} that \textit{LwPosr} has the least number of parameters compared to all other networks in this protocol (to our best knowledge). It has improved the performance over SSR-Net-MD (previous benchmark for least number of parameters) by 0.25 on MAE. TriNet performs quite efficiently on $P_2$ but it has $\sim163.5$ times more parameters ($\sim26 \times 10^6$ as compared to $0.159 \times 10^6$). VGG16 and VGG16+RNN has 870 times more number of parameters than \textit{LwPosr}. Thus, even though these networks have slightly better performance than \textit{LwPosr} on $P_2$, these are difficult to be used on small mobile devices.

\begin{table}
\caption{Results for Protocol $P_2$ using BIWI dataset as 70\% training and 30\% testing data (using only RGB images).}
\label{Tab:results_biwi}
\small
\centering
\scalebox{0.9}{
\begin{tabular}{llllll}
\hline
            & Param $10^6$    & Yaw  & Pitch & Roll & MAE  \\
\hline
\multicolumn{5}{l}{\textbf{RGB-based}}                \\
\hline
DeepHeadPose~\cite{mukherjee2015deep} &-  & 5.67 & 5.18  & -    & -    \\
SSR-Net-MD~\cite{yang2018ssr}  &0.2 & 4.24 & 4.35  & 4.19 & 4.26 \\
VGG16~\cite{gu2017dynamic} &  $\sim$138 & 3.91 & 4.03  & 3.03 & 3.66 \\
FSA-Caps-Fusion~\cite{yang2019fsa} & 1.2 & 2.89 & 4.29  & 3.60 & 3.60 \\
TriNet  &  $\sim$ 26 &2.99 & 3.04 & 2.44 & 2.80 \\
\textit{LwPosr} & \textbf{0.15} & 3.62 & 4.65 & 3.78 & 4.01 \\

\hline
\multicolumn{5}{l}{\textbf{RGB+Depth}}                \\
\hline
DeepHeadPose~\cite{mukherjee2015deep} &-   & 5.32 & 4.76  & -    & -    \\
Martin~\cite{martin2014real}     &-    & 3.6  & 2.5   & 2.6  & 2.9  \\
\hline
\multicolumn{5}{l}{\textbf{RGB+Time}}                 \\
\hline
VGG16+RNN~\cite{gu2017dynamic}   &  $\sim$138+ & 3.14 & 3.48  & 2.6  & 3.07
\end{tabular}}
\vspace{-3mm}

\end{table}
\subsection{Visualization of Results}

\begin{figure*}[htbp!]
    \centering
    \begin{subfigure}[t]{0.49\textwidth}
        \centering
        \includegraphics[height=1.9in,width=2.7in]{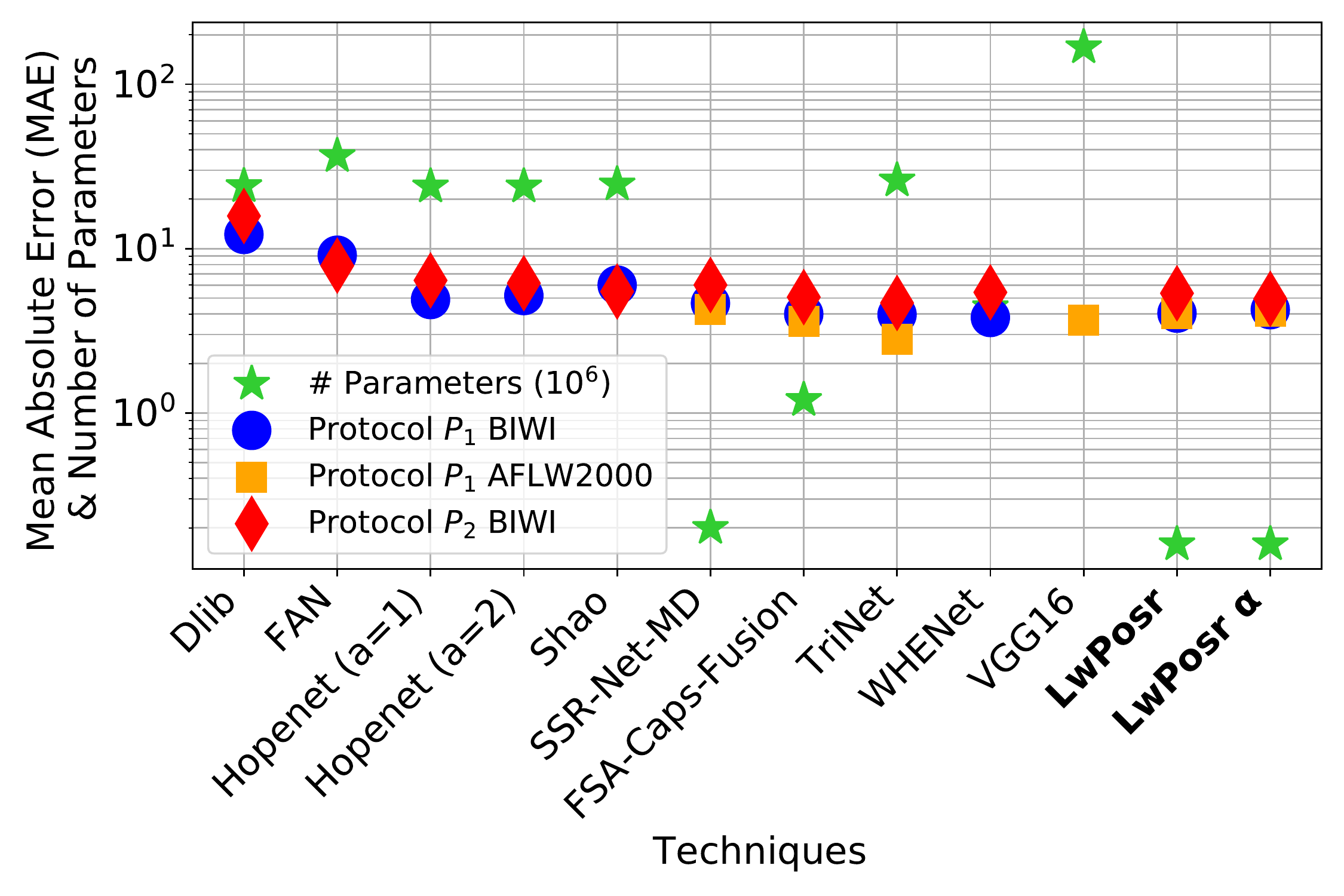}
        \caption{All Compared Techniques}
        \label{Fig:log_grapha}
    \end{subfigure}\hfill 
    \begin{subfigure}[t]{0.49\textwidth}
        \centering
        \includegraphics[height=1.9in,width=2.7in]{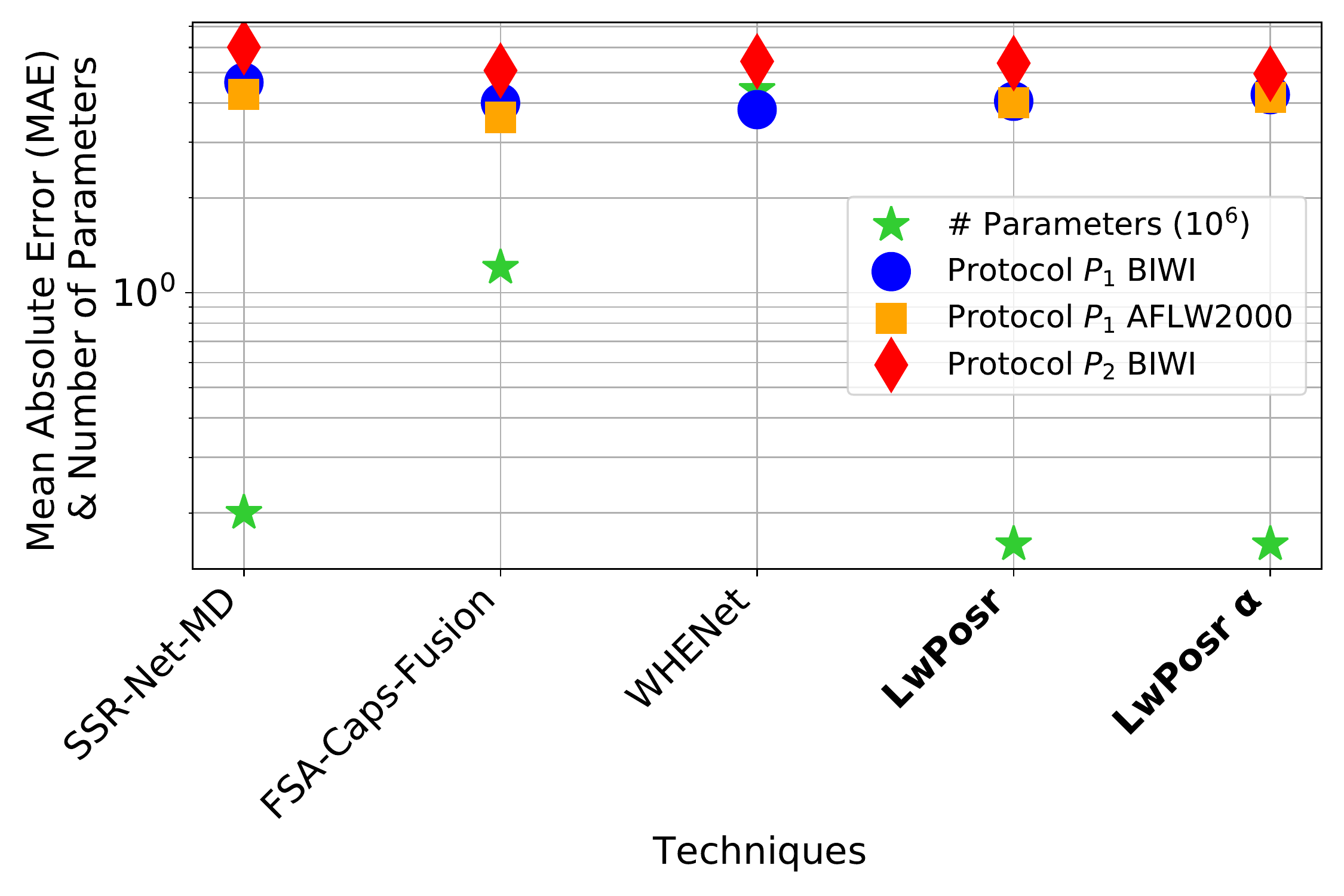}
        \caption{Techniques with Number of Parameters < $5 \times 10^6$}
        \label{Fig:log_graphb}
    \end{subfigure}
    \caption{Visualization of Mean Absolute Error (MAE) and number of parameters for the compared techniques of head pose estimation. It is evident that \textit{LwPosr} has the least number of parameters compared to other techniques. Its performance is also significantly better as compared to most of the networks (surpassing the state-of-the-art memory efficient network, i.e. SSR-Net-MD)}
    \label{Fig:log_graph}
\end{figure*}

Figure \ref{Fig:log_grapha} shows a plot with MAE and number of parameters plotted for the compared HPE techniques. It is evident that \textit{LwPosr} dominates the other networks (TriNet, WHENet, FSA-Caps-Fusion) in parameter number domain and is sub-dominant in MAE. Rest of the networks have more number of parameters and also perform lower than the proposed architecture.

Figure \ref{Fig:log_graphb} shows the networks with number of parameters less than $5 \times 10^6$.  SSR-Net-MD which has the least number of parameters in literature has $0.41 \times 10^6$ more paramters than \textit{LwPosr}. FSA-Caps-Fusion has almost 10 times more number of parameters than \textit{LwPosr}.

\begin{figure*}[htbp!]
\setlength\abovecaptionskip{-0\baselineskip}
\setlength\belowcaptionskip{2pt}
\begin{center}
\includegraphics[height=5cm, width=13.5cm]{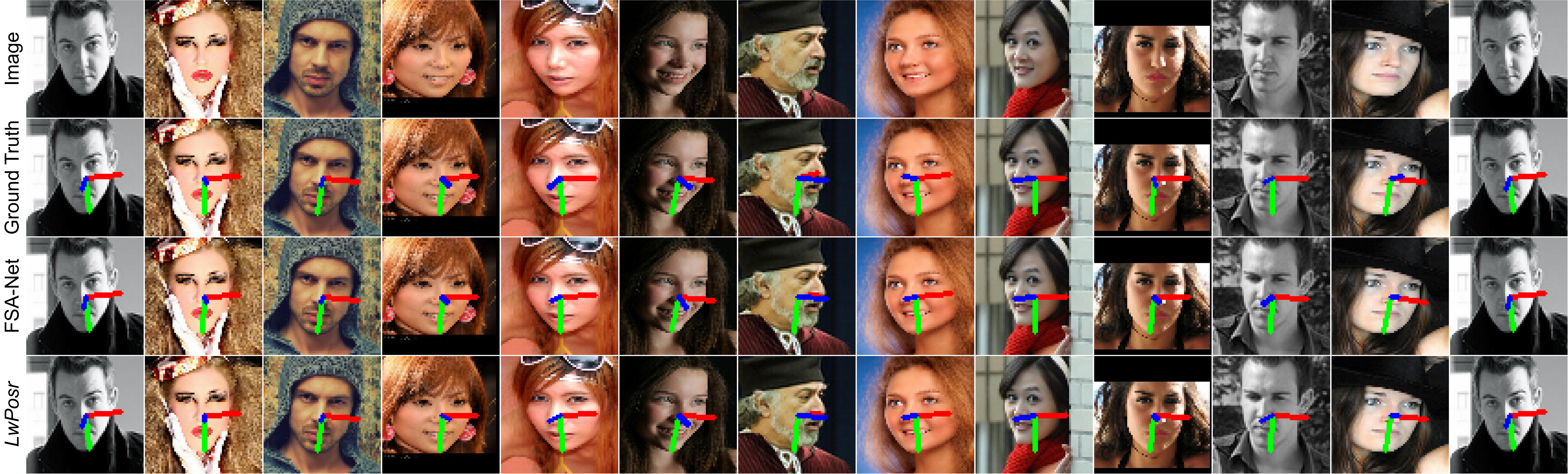}
\caption{Qualitative visualization of the head pose estimation illustration for ground-truth, FSA-Net \cite{yang2019fsa}, \textit{LwPosr}}
 \label{Fig:qualitative_results}
\end{center}
 \vspace{-5mm}
\end{figure*}

Figure \ref{Fig:qualitative_results} shows the qualitative examples of the head pose illustration on the test images of AFLW2000 dataset using protocol $P_1$. The illustrative images in Figure \ref{Fig:qualitative_results} also show the comparison with FSA-Net results. It is seen that \textit{LwPosr} performs equally well or even better on the given examples. For instance, in the first column the blue line is better corresponding to the ground truth as compared to the FSA-Net results. Results are added for BIWI dataset in supplementary material.

\vspace{-1mm}
\subsection{Ablation Study}\label{subsec:ablation}
\vspace{-1mm}

\begin{table*}[hbtp!]
\vspace{-2mm}
\caption{Ablation Study on \textit{LwPosr} network. The ablations are performed on varying the parameters of transformer encoder, changing hyper-parameters, using different weighted mean for final prediction.}
  \label{Tab:ablationtransformer}
  \small
\centering
\begin{tabular}{l|l|llll|llll}
\hline
& & &BIWI &  &  &   & AFLW &  &  \\
\hline
           &Param $10^6$ & Yaw  & Pitch & Roll & MAE & Yaw  & Pitch & Roll & MAE \\
\hline
\multicolumn{5}{l}{\textbf{No. of Encoder Layers in Transformer Encoder}}                \\
\hline
1 & 0.091 & 4.66 & 5.29 & 3.73 & 4.56 & 5.04 & 6.33 & 4.99 & 5.45  \\
2 & 0.126 &4.50 & 5.07 & 3.19 & 4.25 & 4.97 & 6.39 & 5.00 & 5.45  \\
3  & 0.159 & 4.11 & 4.87 & 3.19 & \textbf{4.05} & 4.80 & 6.38 &4.88 & \textbf{5.35} \\
4  & 0.194 & 4.40 & 5.29& 3.33 &4.34 & 4.89 & 6.46 & 5.03 & 5.46 \\
5 & 0.228 & 9.87 & 9.90& 5.96 & 8.57  & 12.39 & 8.71& 9.43 & 10.17 \\
6 & 0.262 &10.15 & 10.21 & 5.94 & 8.77 & 12.24 & 8.75 & 9.27& 10.09 \\
\hline
\multicolumn{5}{l}{\textbf{No. of Heads in Transformer Encoder}}                \\
\hline
1 & 0.159 & 17.62 & 21.06 & 6.97 & 15.22 & 23.25 & 14.47 & 12.53 & 16.75\\
2 & 0.159 & 5.07 & 7.08 & 4.21 & 5.45 & 6.88 & 7.70 & 6.96 & 7.18 \\
4 & 0.159 & 5.28 & 5.95 & 4.37 & 5.2 & 6.55 & 7.80 & 6.68 & 6.99 \\
8  & 0.159 & 4.11 & 4.87 & 3.19 &\textbf{ 4.05} & 4.80 & 6.38 &4.88 &\textbf{ 5.35 }\\
16 & 0.159 & 4.69 & 7.76 & 4.18 & 5.54 & 7.25& 8.35 & 7.52 & 7.80 \\

\hline
\multicolumn{5}{l}{\textbf{Activations in Transformer Encoder}}                 \\
\hline
Relu & 0.159 & 4.11 & 4.87 & 3.19 & \textbf{4.05} & 4.80 & 6.38 & 4.88 & 5.35 \\
Gelu & 0.159 & 4.41 & 5.11 & 3.24 & 4.25 & 4.44 & 6.06 & 4.35  & \textbf{4.95}\\
\hline
\multicolumn{5}{l}{\textbf{Position Embeddings in Transformer Encoder}}                 \\
\hline 
No Embedding & 0.139 & 4.34 & 5.15 & 3.31& 4.26 & 5.35 & 6.39& 5.30 & 5.69\\
Learnable & 0.159 & 4.71& 5.73 & 3.42 & 4.62 & 4.86& 6.38& 4.99 &5.41 \\
Sine  & 0.159 & 4.11 & 4.87 & 3.19 & \textbf{4.05} & 4.80 & 6.38 & 4.88 & \textbf{5.35 }\\

\hline
\multicolumn{5}{l}{\textbf{Learning Rate}}                 \\
\hline
0.01 & 0.159 & 10.27& 9.57 & 5.67 & 8.50 & 12.30 & 8.77 & 9.41 &10.16 \\ 
0.001 & 0.159 & 4.11 & 4.87 & 3.19 & \textbf{4.05} & 4.80 & 6.38 & 4.88 & \textbf{5.35 }\\
0.0001 & 0.159 & 5.16 & 6.43 & 3.87 & 5.15 & 6.48 &7.57 &6.52 &6.86\\
\hline
\multicolumn{5}{l}{\textbf{Weights in the Weighted Mean}}                \\
\hline
$[0.5, 0.5, 2]/3$   & 0.159 & 4.11 & 4.87 & 3.19 & \textbf{4.05} & 4.80 & 6.38 &  4.88 & 5.35 \\
$[2, 0.5, 0.5]/3$ & 0.159 &4.26 & 5.10 & 3.40 & 4.25 & 4.85 & 6.21 & 4.79& \textbf{5.28} \\
$[1, 1, 1]/3$ & 0.159 & 4.50 & 5.46 & 3.08 & 4.34 & 4.93 & 6.33 & 5.10 & 5.45\\

Learnable & 0.159 & 4.50 & 5.37 & 3.20 & 4.35& 4.74 &6.63 & 5.18 & 5.52\\
\hline

\multicolumn{5}{l}{\textbf{Loss Functions }}                \\
\hline
MAE &0.159 & 4.11 & 4.87 & 3.19 & \textbf{4.05} & 4.80 & 6.38 &4.88 & 5.35\\
MAE + Ortho loss &0.159 & 4.29& 5.24 & 3.38 & 4.30& 4.84 & 6.31& 4.77 & \textbf{5.31}\\
\hline
\multicolumn{5}{l}{\textbf{No. of stages in \textit{LwPosr}}}                \\
\hline
3  & 0.159 & 4.11 & 4.87 & 3.19 & \textbf{4.05} & 4.80 & 6.38 &4.88 & 5.35 \\
4 & 0.221 & 4.69  & 5.63 & 3.51 & 4.61 & 5.22 & 6.62 & 4.97 & \textbf{5.27}  \\
\hline

\end{tabular}
\vspace{0mm}
\end{table*}
All the results corresponding to ablations performed using Protocol $P_1$ are concisely described in Table \ref{Tab:ablationtransformer} (parameter being studied is varied and other parameters are kept constant to 3 encoder layers, 4 heads, relu activation function, sine embedding, 0.001 learning rate, $[0.5, 0.5, 2]/3$ weights in the weighted mean, MAE loss, 3 stages).

\textbf{Weights in the Weighted Mean:} Eq. \ref{eq:combine} shows that the outputs from all the mixture models have own individual head pose predictions. These predictions are combined by weighted means. The experiments performed for $\alpha_1, \alpha_2, \alpha_3$  for stages 1, 2, and 3, respectively, having sets of values [0.5, 0.5, 2]/3, [2, 0.5, 0.5]/3, and [1, 1, 1]/3 show that the set of values [0.5, 0.5, 2]/3 performs better than other sets. It is intuitive that the last stage rectifies the features learnt by the initial stages. So, the performance is better when more weight is given to the last stage. Investigation is also performed on learnable set (where $\alpha_1, \alpha_2, \alpha_3$ are learnable parameters). But it performs worse than the fixed sets.

\textbf{Learning Rate:} Initial learning rates are investigated for values of 0.01, 0.001, 0.0001. It is evident from Table \ref{Tab:ablationtransformer} that training gets stuck in local mimina with the learning rate of 0.01. The learning rate of 0.001 performs better than the other two. 

\textbf{Activation Function in Transformer Encoder:} Two activation functions are checked for in transformer encoder, i.e., Relu and Gelu functions. For BIWI dataset, Relu shows to perform better than Gelu function and for AFLW2000 dataset Gelu performs better than Relu function in the proposed network.

\textbf{Position Embeddings:} The 2D spatial structure is learnt using  position embedding in transformer. Three tests are performed: (1) with no position embedding, (2) learnable embedding, and (3) sine embedding. Intuitively, network with no position embedding performs the worst. It makes the transformer encoder permutation-invariant. Sine embedding performs better than learnable embedding for \textit{LwPosr} (details of sine-embedding are mentioned in Supplementary material).

\textbf{Size Scaling of Transformer Encoder:} Number of encoder layers are varied to check their effect in \textit{LwPosr} on the HPE. The experiments are performed for one to six encoder layers. It is seen that the optimized number of encoder layers is three. If the number of encoders is increased further than three, then the MAE increases. 

\textbf{Number of Heads in Transformer Encoder:} Experiment by varying the number of heads in transformer encoder are performed. It is seen that the performance increases by increasing the number of heads until eight heads and then starts declining.

\textbf{Loss Functions:} Orthogonality (Ortho) loss as used in \cite{cao2021vector} is investigated by using Ortho loss and MAE together for optimization. It is seen in the experiments that using Ortho loss in addition to MAE does not improve the performance in the network (details on Ortho loss in supplementary material).
\vspace{-2mm}

\section{Different Stages in \textit{LwPosr} Network}
We implemented 4 stages in \textit{LwPosr}, where $Q=4$ instead of $Q=3$ (as mentioned in section \ref{sec:method}). Table~\ref{Tab:ablationtransformer} illustrates the number of parameters and results for the corresponding stages. It shows that increasing the stages increases the number of parameters which have to be learnt. Hence, the network size increases. The quantitative results on average do not change. So, network with three stages is more optimized in terms of number of parameters and evaluation results.

\vspace{0mm}
\section{Conclusion}
\vspace{-2mm}
This paper presented a novel architecture \emph{LwPosr} for head pose predictions. In contrast to previous work, this model is able to predict poses efficiently by taking a very small number of parameters and memory. A key insight is that transformer encoder helps in learning the global information. Using transformer encoder in combination with the depthwise separable convolutions gets the best of both entities and helps to learn spatial as well as global features. Furthermore, transformer encoders take less number of parameters, hence small network size. It is demonstrated that proposed model outperforms previous memory efficient approaches on 1) lowering mean absolute error and 2) decreasing the number of parameters. It is believed that in future crucial concepts of the presented work can be used in other domains and tasks that have a similar problem statement.
\vspace{-3mm}

{\small
\bibliographystyle{ieee_fullname}
\bibliography{main}
}

\end{document}